\documentclass[]{imsart}
\usepackage[english]{babel}
\usepackage{amsbsy,amssymb}
\usepackage{booktabs}
\usepackage{graphpap,latexsym,epsf}
\usepackage{graphicx}
\usepackage{tabularx}
\usepackage{dsfont}
\usepackage{bbm}
\usepackage{bm}
\usepackage{mathtools}
\usepackage{stmaryrd}
\usepackage{float}
\usepackage{afterpage}
\usepackage{emptypage}
\usepackage{rotating}  
\usepackage{fancyhdr}  
\usepackage{enumitem}
\usepackage{colortbl}
\usepackage[table]{xcolor} 
\usepackage{wrapfig} 

\graphicspath{ {picture/} }

\RequirePackage[OT1]{fontenc}
\RequirePackage{amsthm,amsmath,natbib}
\RequirePackage[colorlinks,citecolor=blue,urlcolor=blue]{hyperref}

\startlocaldefs
\numberwithin{equation}{section}
\theoremstyle{plain}
\newtheorem{thm}{Theorem}[section]
\newtheorem{proposition}[thm]{Proposition}
\newtheorem*{proposition*}{Proposition}
\theoremstyle{remark}

\endlocaldefs

\begin{document}

\begin{frontmatter}
\title{Conformal Prediction: a Unified Review of Theory and New Challenges}
\runtitle{Conformal Prediction: a Unified Review}

\begin{aug}

\author{\fnms{Matteo} \snm{Fontana}\thanksref{a,b,c}\ead[label=e2,mark]{matteo.fontana@ec.europa.eu}}
\author{\fnms{Gianluca} \snm{Zeni}\thanksref{a}}
\and
\author{\fnms{Simone} \snm{Vantini}\thanksref{a}}

\address[a]{MOX-Department of Mathematics, Politecnico di Milano, Italy}
\address[b]{Department of Management, Economics and Industrial Engineering, Politecnico di Milano, Italy}
\address[c]{now at European Commission, Joint Research Centre (JRC), Ispra, Italy, email: matteo.fontana@ec.europa.eu}

\runauthor{M. Fontana, G. Zeni and S. Vantini}
\end{aug}

\begin{abstract}

In this work we provide a review of basic ideas and novel developments about Conformal Prediction --- an innovative distribution-free, non-parametric forecasting method, based on minimal assumptions --- that is able to yield in a very straightforward way prediction sets that are valid in a statistical sense also in the finite sample case.
The discussion provided in the paper covers the theoretical underpinnings of Conformal Prediction, and then proceeds to list the more advanced developments and adaptations of the original idea.

\end{abstract}


\end{frontmatter}

\section{Introduction}

In the the latest 25 years, a new method of prediction with confidence, called Conformal Prediction (CP), was introduced and developed.
CP allows to produce prediction sets with the guaranteed error rate, exclusively under the simple i.i.d. assumption of the sample. 
Reliable estimation of prediction confidence is a significant challenge in both machine learning and statistics, and the promising results generated by CP have generated further extensions of the original conformal framework.
The increasing amount of real-world problems where robust predictions are needed is testified by the multitude of scientific works where CP is used.

In a nutshell, conformal prediction uses past experience in order to determine precise levels of confidence in new predictions. 
Using \cite{gammerman1998learning}'s words in the very first work on the topic, it is ``a practical measure of the evidence found in support of that prediction''. 
In order to do this, it estimates how ``unusual'' a potential example looks with respect to the previous ones. Prediction regions are generated plainly by including the examples that have quite ordinary values, or better those ones that are not very unlikely. 
Conformal algorithms are proven to be always valid: the actual confidence level is the nominal one, without requiring any specific assumption on the distribution of the data except for the i.i.d. assumption. 
There are many conformal predictors for any particular prediction problem, whether it is a classification problem or a regression problem. Indeed, we can construct a conformal predictor from any method for scoring the similarity (conformity, as it is called) of a new example with respect to the old ones. 
For this reason, it can be used with any statistical and machine learning algorithm.
Efficient performances let to understand the growing interest on the topic over the last few years.

The cornerstone of the related literature is "Algorithmic learning in a random world", published in $2005$, where Vladimir Vovk and coauthors explains thoroughly all the theoretical fundamentals of CP 
There is only another, more recent, work that gives an overview on the topic, namely the book Conformal prediction for reliable machine learning, edited by \citeauthor{book:2014} (\citeyear{book:2014}). 
The mentioned book addresses primarily applied researchers, showing them the practical results that can be achieved and allowing them to embrace the possibilities CP is able to give.
Therefore, its focus is almost totally on adaptations of conformal methods and the connected real-world applications.  

In the latest years, an extensive research effort has been pursued with the aim of extending the framework, and several novel findings have been made. We have no knowledge of in-depth publications that aim to capture these developments and give a picture of recent theoretical breakthroughs.  
Moreover, there is a great deal of inconsistencies in the extensive literature that has been developed (e.g. regarding notation.), so the need for such an up-to-date review is then evident, and the aim of this work is to address this need of comprehensiveness and homogeneity.

As in recent papers, our discussion is focused on CP in the batch mode. Nonetheless, properties and results concerning the online setting, which tended to be the focus of the earliest works on the subject, are not omitted.

The paper is divided into two parts: Part~\ref{part:one} gives an introduction to CP for non-specialists, explaining the main algorithms, describing their scope and also their limitations, while Part~\ref{part:two} discusses more advanced methods and developments.
Section~\ref{sec:basics} introduces comprehensively the original version of conformal algorithm, and let the reader familiarize with the topic and the notation.
In Section~\ref{sec:objandlab}, a simple generalization is introduced: each example is provided with a vector of covariates, like in classification or regression problems, which are tackled in the two related subsections. 
Section 1 of the Supplementary Material \citep{zeni_supp} shows a comparison between CP and alternative ways of producing confidence predictions, namely the Bayesian framework and the statistical learning theory, and how CP is able to overcome their weak points. 
Moreover, we refer to an important result concerning the optimality of conformal predictors among valid predictors (In Section 1.1 of the Supplementary Material \citep{zeni_supp}). 
Section~\ref{sec:online} deals with the online framework, where examples arrive one by one and so predictions are based on an accumulating data set. 

In Part~\ref{part:two}, we focus on three important methodological themes. 
The first one is the concept of statistical validity: Section~\ref{sec:validity} is entirely devoted to this subject and introduces a class of conformal methods, namely Mondrian conformal predictors, suitable to gain object conditional validity in partial sense.
Secondly, computational problems, and a different approach to conformal prediction --- the inductive inference --- to overcome the transductive nature of the basic algorithm (Section~\ref{sec:inductive}).
Even in the inductive formulation, the application of conformal prediction in the case of regression is still complicated, but there are ways to face this problem (Section 2 of the Supplementary Material \citep{zeni_supp}).  
Lastly, what is historically called as "the randomness assumption": conformal prediction is valid if examples are sampled independently from a fixed but unknown probability distribution. 
It actually works also under the slightly weaker assumption that examples are probabilistically exchangeable, and under other online compression models, as the widely used Gaussian linear model (Section 3 of the Supplementary Material \citep{zeni_supp})) . 

The last section (Section~\ref{sec:other}) addresses interesting directions of further development and research. 
We describe extensions of the framework that improve the interpretability and applicability of conformal inference.
CP has been applied to a variety of applied tasks and problems. For this reason it is not possible to refer to all of them here: the interested reader can find an exhaustive list in \cite{book:2014}.

\section{Foundations of Conformal Prediction}
\label{part:one}

\subsection{Conformal Predictors}\label{sec:basics}
We will now show how the basic version of CP works. In the basic setting, successive values $ z_1, \, z_2, \, z_3, \dots \in \mathbf{Z}$, called \textit{examples}\/, are observed.
$\mathbf{Z}$ is a measurable space, called the examples space. 
We also assume that $\mathbf{Z}$ contains more than one element, and that each singleton is measurable. 
Before the $(n+1)$th value $z_{n+1}$ is announced, the training set\footnote{From a mathematical point of view it is a sequence, not a set.} consists of $(z_1, \dots, z_n)$ and our goal is to predict the new example.

\begin{sloppypar}
To be precise, we are concerned with a prediction algorithm that outputs a set of elements of $\mathbf{Z}$, implicitly meant to contain $z_{n+1}$.
Formally, a \textit{prediction set\/} is a measurable function $\gamma$ that maps a sequence ${(z_1,\dots,z_n) \in \mathbf{Z}^n}$ to a set $\gamma(z_1,\dots,z_n) \subseteq \mathbf{Z}$, where the measurability condition reads as follow: the set $
\{ (z_1,\dots,z_{n+1}) : z_{n+1} \in \gamma(z_1,\dots,z_n) \} $ 
is measurable in $\mathbf{Z}^{n+1}$.
A trade-off between reliability and informativeness has to be faced by the algorithm while giving as output the prediction sets. Indeed giving as a prediction set the whole examples space $\mathbf{Z}$ is not appealing nor useful: it is absolutely reliable but not informative.
\end{sloppypar}

Rather than a single set predictor, we are going to deal with nested families of set predictors depending on a parameter $\alpha \in [0,1]$, the \textit{significance level}\/ or target miscoverage level, reflecting the required reliability of the prediction. 
The smaller $\alpha$ is, the bigger the reliability in our guess. So, the quantity $1 - \alpha$ is usually called the confidence level.
As a consequence, we define a \textit{confidence predictor\/} to be a nested family of set predictors $(\gamma^{\alpha})$, such that, given $\alpha_1$, $\alpha_2$ and $0 \leq \alpha_1 \leq \alpha_2 \leq 1$,
\begin{equation}
    \gamma^{\alpha_1}(z_1,\dots,z_n) \supseteq \gamma^{\alpha_2}(z_1,\dots,z_n).
\end{equation}

Confidence predictors from old examples alone, without knowing anything else about them, may seem relatively uninteresting. 
But the simplicity of the setting makes it advantageous to explain and understand the rationale of the conformal algorithm, and, as we will see, it is then straightforward to take into account also features related to the examples. 

In the greatest part of the literature concerning conformal prediction, and especially in the works of Vovk and coauthors, the symbol $\varepsilon$ stands for the significance level. 
Nonetheless, we prefer to adopt the symbol $\alpha$, as in \cite{lei2013distribution}, to be faithful to the statistical tradition and its classical notation. 
For the same reason, we want to predict the $(n+1)$th example, relying on the previous experience given by $(z_1, \dots, z_n)$, still like \citeauthor{paper:lei2017} and conversely to \citeauthor{book:alrw}. The latter is interested in the $n$th value given the previous $(n-1)$ ones. 

\subsubsection{The Randomness Assumption}
We will make two main kinds of assumptions about the way examples are generated. 
The standard assumption is the usual i.i.d. one commonly employed in the statistical setting: the examples we observe are sampled independently from some unknown probability distribution $P$ on $\mathbf{Z}$. 
Equivalently, the infinite sequence $z_1, z_2, \dots$ is drawn from the power probability distribution $P^{\infty}$ in $\mathbf{Z}^{\infty}$.

Under the exchangeability assumption, instead, the sequence $(z_1, \dots, z_n)$ is generated from a probability distribution that is exchangeable: for any permutation $\pi$ of the set $\{1, \dots, n\}$, the joint probability distribution of the permuted sequence $(z_{\pi(1)}, \dots, z_{\pi(n)})$ is the same as the distribution of the original sequence. 
In an identical way, the $n!$ different orderings are equally likely. 
It is possible to extend the definition of exchangeability to the case of an infinite sequence of variables: $z_1, z_2, \dots$ are exchangeable if $z_1, \dots, z_N$ are exchangeable for every $N$.

Exchangeability implies that variables have the same distribution. On the other hand, exchangeable variables need not to be independent. It is immediately evident how the exchangeability assumption is weaker than the randomness one. 
As we will see in Section~\ref{sec:online}, in the online setting the difference between the two assumptions almost disappears.
For further discussions about exchangeability, including various definitions, a game-theoretic approach and a law of large numbers, refer to Section~3 of \cite{paper:tutorial}.

The randomness assumption is a standard assumption in machine learning. Conformal prediction, however, usually requires only the sequence $(z_1, \dots, z_n)$ to be exchangeable. 
In addition, other models which do not require exchangeability can also use conformal prediction (Section 3 of the Supplementary Material \citep{}).

\subsubsection{Bags and Nonconformity Measures}

First, the concept of a nonconformity (or strangeness) measure has to be introduced. In few words, it estimates how unusual an example looks with respect to the previous ones. 
The order in which old examples $(z_1, \dots, z_n)$ appear should not make any difference. To underline this point, we will use the term \textit{bag\/} (in short, $B$) and the notation $\Lbag z_1, \dots, z_{n} \Rbag.$.
A bag is defined exactly as a multiset. Therefore, $\Lbag z_1, \dots, z_{n} \Rbag$ is the bag we get from $(z_1, \dots, z_n)$ when we ignore which value comes first, which second, and so on. 
To increase clarity, we specifiy that the main difference between a bag/multiset and a set is that an element in a bag/multiset can be repeated.

As mentioned, a \textit{nonconformity measure\/} (NCM) $A\!:\mathbf{Z}^n \times \mathbf{Z} \rightarrow \mathbb{R}$ is a way of scoring how different an example $z \in \mathbb{R}$ is from a bag $B \in \mathbf{Z}^n \times \mathbf{Z}$.
There is not just one nonconformity measure. For instance, once the sequence of old examples $(z_1, \dots, z_n)$ is at hand, a natural choice is to take the average as the simple predictor of the new example, and then compute the nonconformity score as the absolute value of the difference from the average. In more general terms, the distance from the central tendency of the bag might be considered.
As pointed out in \cite{book:alrw}, whether a particular function $A$ is an appropriate way of measuring nonconformity will always be open to discussion, as it greatly depends on contextual factors.

We have previously remarked that $\alpha$ represents our significance level. Now, for a given nonconformity measure $A$, we set $R = A(B,z)$ to stand for the nonconformity score --- where $R$ is related in a certain way to the word ``residual''. 
On the contrary, most of the literature uses $\varepsilon$ and $\alpha = A(B,z)$, respectively. We still prefer \citeauthor{paper:lei2017}'s notation.

Instead of a nonconformity measure, a conformity one might be chosen. The line of reasoning does not change at all: we could compute the scores and resume to the first framework just by changing the sign, or computing the inverse. However, conformity measures are not a common choice.

\subsubsection{Conformal Prediction}

The idea behind conformal methods is extremely simple. 
Consider $n+1$ exchangeable observations of a scalar random variable, let's say $u_1, \dots, u_{n+1}$. 
In the absence of ties, the rank of the $n+1$th observation $u_{n+1}$ among $u_1, \dots, u_{n+1}$ is uniformly distributed over the set $\{1,\dots,n+1\},$ due to exchangeability.

Back to the nonconformity framework, under the assumption that the $z_i$ are exchangeable, we define, for a given $z \in \mathbf{Z}$:
\begin{equation} \label{eq:pvalue}
    p_z \coloneqq \dfrac{ \lvert \{ \, i = 1, \dots, n+1 : R_i \geq R_{n+1}\} \rvert }{n+1}
\end{equation}
where
\begin{equation}\label{eq:residualcp}
\begin{aligned}
    R_i & = A \, ( \, \Lbag z_1, \dots, z_{i-1}, z_{i+1}, \dots, z_n, z \Rbag \, , z_i)  \\ 
    & = A \, ( \, \Lbag z_1, \dots, z_n, z \Rbag \, \setminus \Lbag z_i \Rbag \, , z_i) \qquad \quad \forall \: i = 1, \dots, n 
\end{aligned}
\end{equation}
and
\begin{equation}
    R_{n+1} \coloneqq A \, ( \, \Lbag z_1, \dots, z_{n} \Rbag \, , z).
\end{equation} 
 It is straightforward that $p_z$ stands for the fraction of examples that are as or more different from the all the others than $z$ actually is. This fraction, which lies between $\frac{1}{n+1}$ and $1$, is defined as the \textit{p-value\/} for $z$. 
If $p_z$ is small, then $z$ is very nonconforming with respect to the past experience, represented by $(z_1, \dots, z_n)$. On the contrary, if large, then $z$ is very conforming and likely to appear as the next observation. Hence, it is reasonable to include it in the prediction set.

As a result, we define the prediction set $\gamma^{\alpha}(z_1,\dots,z_{n})$ by including all the $z$s that conform with the previous examples. 
In a formula, 
$ \gamma^{\alpha}(z_1,\dots,z_{n}) \coloneqq \{ z \in \mathbf{Z} : p_z > \alpha \}. $
To summarize, the algorithm tells us to form a prediction region consisting of all the $z$s that are not among the fraction $\alpha$ most out of place with respect to the bag of old examples. 
\cite{paper:tutorial} give also a clear interpretation of $\gamma^{\alpha}(z_1,\dots,z_{n})$ as an application of the Neyman-Pearson theory for hypothesis testing and confidence intervals. 

\subsubsection{Validity and Efficiency}
The two main indicators of how good confidence predictors behave are validity and efficiency, respectively an index of reliability and informativeness. 
A set predictor $\gamma$ is \textit{exactly valid\/} at a significance level $\alpha \in [0,1]$, if the probability of making an error --- namely the event $z_{n+1} \notin \gamma^{\alpha}$ --- is $\alpha$, under any probability distribution on $\mathbf{Z}^{n+1}$.  If the probability does not exceed $\alpha$, under the same conditions, a set predictor is defined as conservatively valid. 
If the properties hold at each of the significance level $\alpha$, the confidence predictor $(\gamma^{\alpha} : \alpha \in [0,1])$ is respectively valid and conservatively valid.
The following result, concerning conformal prediction, holds \citep{book:alrw}:
\begin{proposition}\label{prop:validity}
    Under the exchangeability assumption, the probability of error, $z_{n+1} \notin \gamma^{\alpha}(z_1,\dots,z_{n})$, will not exceed $\alpha$, for any $\alpha$ and any conformal predictor $\gamma$.
\end{proposition}
In an intuitive way, due to exchangeability, the distribution of $(z_1, \dots, z_{n+1})$ and so the distribution of the nonconformity scores $(R_1, \dots, R_{n+1})$ are invariant under permutations; in particular, all permutations are equiprobable.
This simple concept is the bulk of the proof and the key of conformal methods. 

From a practical point of view, the conservativeness of the validity is often not ideal, especially when $n$ is large, and so we get long-run frequency of errors very close to $\alpha$. 
From a theoretical prospective, \cite{paper:lei2017} indeed prove, under minimal assumptions on the residuals, that conformal prediction intervals are accurate, meaning that they do not substantially over-cover. 

A conformal predictor is always conservatively valid. Is it possible to achieve exact validity by introducting a randomized component into the algorithm.
The \textit{smoothed conformal predictor\/} is defined in the same way as before, except that the p-values (\ref{eq:pvalue}) are replaced by the smoothed p-values:
\begin{equation}
    p_z \coloneqq \dfrac{ \lvert \{ \, i : R_i > R_{n+1}\} \rvert + \tau \, \lvert \{ \, i : R_i = R_{n+1}\} \rvert }{n+1}, 
\end{equation}
where the \textit{tie-breaking\/} random variable $\tau$ is uniformly distributed on $ [0, 1]$ ($\tau$ can be the same for all $z$s). 
For a smoothed conformal predictor, as wished, the probability of a prediction error is exactly $\alpha$ (\cite{book:alrw}, Proposition 2.4).

Alongside validity, prediction algorithms should be efficient too, that is to say, the uncertainty related to predictions should be as small as possible. 
Validity is the priority: without it, the meaning of predictive regions is lost, and it becomes easy to achieve the best possible performance.
Without constraints, in fact, the trivial  $\gamma^{\alpha}(z_1,\dots,z_{n-1}) \coloneqq \emptyset$ is the most efficient one. 
Efficiency may appear as a vague notion, but in any case it can be meaningful only if we impose some restrictions on the predictors that we consider.

Among the main problems solved by Machine Learning and Statistics we can find two types of problems: classification, when predictions deal with a small finite set (often binary), and regression, when instead the real line is considered.  
In classification problems, two criteria for efficiency have been used most often in literature.
One criterion takes account of whether the prediction is a single class (the most efficient case), a multiple set of classes, or the entire set of possible classes (the most inefficient case) at a given significance level $\alpha$. 
Alternatively, the confidence and \textit{credibility\/} of the prediction --- which do not depend on the choice of a significance level $\alpha$ --- are considered.
The former is the greatest $1-\alpha$ for which $\gamma^{\alpha}$ is a single label, while the latter, helpful to avoid overconfidence when the object $x$ is unusual, is the largest $\alpha$ for which the prediction set is empty. 
\cite{vovk2016criteria} show several other criteria, giving a detailed depiction of the framework. 
In regression problems instead, the prediction set is often an interval of values, and a natural measure of efficiency of such a prediction is simply the length of the interval. The smaller it is, the better its performance.

We will be looking for the most efficient confidence predictors in the class of valid, or in an equivalent term well-calibrated, confidence predictors; different notions of validity (including conditional validity, examined in Section~\ref{sec:validity}) and different formalisations of the notion of efficiency will lead to different solutions to the problem.

\subsection{Objects and Labels}\label{sec:objandlab}

In this section, we introduce a generalization of the basic CP setting.
A sequence of successive examples $z_1, z_2, z_3, \dots$ is still observed, but each example consists of an \textit{object\/} $x_i$ and its \textit{label\/} $y_i$, i.e $z_i = (x_i, y_i)$.
The objects are elements of a measurable space $\mathbf{X}$ called the object space, and the labels of a measurable space $\mathbf{Y}$ called the label space (both in the classification and the regression contexts). As before, we take for granted that $\lvert \mathbf{Y} \rvert > 1$.
In a more compact way, let $z_i$ stand for $(x_i, y_i)$, and $\mathbf{Z} \coloneqq \mathbf{X} \times \mathbf{Y}$ be the example space. 

At the $(n+1)$th trial, the object $x_{n+1}$ is given, and we are interested in predicting its label $y_{n+1}$.
The general scheme of reasoning is unchanged. Under the randomness assumption, examples, i.e. $(x_i, y_i)$ couples, are assumed to be i.i.d. First, we need to choose a nonconformity measure in order to compute nonconformity scores. Then, p-values are computed, too. Last, the prediction set $\varGamma^{\alpha}$ turns out to be defined as follow:
\begin{equation}\label{eq:Gamma}
\begin{aligned}
    \varGamma^{\alpha} (z_1,\dots,z_{n},x_{n+1}) &\coloneqq \{ y : (x_{n+1}, y) \in \varGamma^{\alpha} (z_1,\dots,z_{n}) \} \\
    & = \{ y \in \mathbf{Y} : p^{(x_{n+1},y)} > \alpha \}
\end{aligned}
\end{equation}

In most cases, the way to proceed, when defining how much a new example conforms with the bag $B$ of old examples, is relying on a simple predictor $f$. The only condition to hold is that $f$ must be invariant to permutations in its arguments ---  equivalently, the output does not depend on the order in which they are presented. 
The method $f$ defines a prediction rule. It is natural then to measure the nonconformity of $z$ by looking at the deviation of the predicted label $ \hat{y}_i =  f_{\,\Lbag z_1, \dots, z_{n} \Rbag} (x_i)$
from the true one. For instance, in regression problems, we can just take the absolute value of the difference between $\hat{y}_i$ and $y_i$. 
That's exactly what we have suggested in the previous (unstructured) case (Section~\ref{sec:basics}), when we proposed to take the mean or the median as the simple predictor for the next observation.
Nevertheless, there are instances in which other NCMs may be more powerful and/or useful: a notable example is Multi-Level Conformal Clustering, and the use of the k-NN distance \citep{nouretdinov2020MLCC} or the use of density estimators in \cite{lei2015conformal}.

Following these steps any simple predictor, combined with a suitable measure of deviation of $\hat{y}_i$ from $y_i$, leads to a nonconformity measure and, therefore, to a conformal predictor. 
The algorithm will always produce valid nested prediction regions. But the prediction regions will be efficient (i.e. small) only if $A(B, z)$ measures well how different $z$ is from the examples in $B$. And consequently only if the underlying algorithm is appropriate. 
Conformal prediction ends up to be a powerful meta-algorithm, created on top of any point predictor --- very powerful but yet extremely simple in its rationale. 

A useful remark in \cite{paper:tutorial} points out that the prediction regions produced by the conformal algorithm do not change when the nonconformity measure $A$ is transformed monotonically. 
For instance, if A is positive, choosing $A$ or its square $A^2$ will make no difference. While comparing the scores to compute $p_z$, indeed, the interest is on the relative values and their reciprocal position --- whether one is bigger than another or not, but not on the single absolute values. 
As a result, the choice of the deviation measure is relatively unimportant. In this framework, the important step in determining the nonconformity measure in this setting is choosing the point predictor $f$.

\subsubsection{Classification}\label{subsec:classification}

In the broader literature, CP has been proposed and implemented with different nonconformity measures for classification --- i.e, when $\lvert \mathbf{Y} \rvert < \infty $. 
As an illustration, given the sequence of old examples $(x_1, y_1), \dots, (x_n, y_n)$ representing past experience, nonconformity scores $R_i$ can be computed as follow:
\begin{equation}\label{eq:knnscore}
    R_i \coloneqq \frac{\min_{\,j = 1, \dots, n \,:\, j \neq i \, \& \, y_j = y_i} \, \Delta (x_i, x_j)}{\min_{\,j = 1, \dots, n \, :\, j \neq i} \, \, \Delta (x_i, x_j)}
\end{equation}
where $\Delta$ is a metric on $\mathbf{X}$, usually the Euclidean distance in an Euclidean setting. 
The rationale behind the scores (\ref{eq:knnscore}) --- in the spirit of the $1$-nearest neighbor algorithm --- is that an example is considered nonconforming to the sequence if it is close to examples labeled in a different way and far from the ones with the same label. 
In a different way, we could use a nonconformity measure that takes account of the average values for the different labels, and the score $R_i$ is simply the distance to the average of its label. 

As an alternative, nonconformity scores can be extracted from the support vector machines trained on $(z_1, \dots, z_n)$. 
We consider in particular the case of binary classification, as the first works actually did to face this problem  \citep{gammerman1998learning,saunders1999transduction}, but there are also ways to adapt it to solve multi-label classification problems \citep{book:2014}.
A plain approach is defining nonconformity scores as the values of the Lagrange multipliers, that stand somehow for the margins of the probability estimating model. If an example's true class is not clearly separable from other classes, then its score $R_i$ is higher and, as desired, we tend to classify it as strange.

Another example of nonconformity measure for classification problems is \cite{devetyarov2010prediction}, who rely on random forests. For instance, a random forest is constructed from the data sequence, and the conformity score of an example $z_i$ is just equal to the percentage of correct predictions for its features $x_i$ given by decision trees.

\subsubsection{Regression}\label{subsec:regression}
In regression problems, a very natural nonconformity measure is: 
\begin{equation}\label{eq:residgeneral}
    R_i \coloneqq \Delta (y_i, f(x_i))
\end{equation} 
where $\Delta$ is a measure of difference between two labels (usually a metric) and $f$ is a prediction rule (for predicting the label given the object) trained on the set $(z_1, \dots, z_{n})$.

It is evident how there is a fundamental problem in implementing conformal prediction for regression tasks: to form the prediction set (\ref{eq:Gamma}), examining each potential label $y$ is needed. 
Nonetheless, there is often a feasible way to compute (\ref{eq:Gamma}) which does not require to examine infinitely many cases; in particular, this happens when the underlying simple predictor is ridge regression or nearest neighbors regression.  
We are going to provide a sketch of how it works, to give an idea of the way used to circumvent the unfeasible brute-force, testing-all approach.
Besides, a slightly different approach to conformal prediction has been developed and carried on to overcome this difficulty (Section~\ref{sec:inductive}, Section 2 of the Supplementary Material \citep{zeni_supp}). 

In the case where $\Delta (y_i, \hat{y}_i) = \lvert y_i - \hat{y}_i \rvert $ and $f$ is the ridge regression procedure, the conformal predictor is called the ridge regression confidence machine (RRCM). 
The initial attempts to apply conformal prediction in the case of regression involve exactly ridge regression (\cite{melluish1999transduction}, and soon after, in a more in-depth version, \cite{nouretdinov2001ridge}).
Suppose that objects are vectors consisting of $d$ attributes in a Euclidean space, say $\mathbf{X} \subseteq \mathbb{R}^d $, and let $\lambda$\footnote{$\lambda$ can be selected in a non-adaptive fashion, or an in adaptive one, exploiting techniques such as cross-validation.} be the non-negative constant called the ridge parameter --- least squares is the special case corresponding to $\lambda = 0$
The explicit representation, in matrix form, of this nonconformity measure is:
\begin{equation}\label{eq:residualrrcm}
     R_i \coloneqq \lvert y_i - x_i'(X'X + \lambda I)^{-1} X'Y \rvert,
\end{equation}
where $X$ is the $n+1 \times d$ object matrix whose rows are $x'_1, x'_2, \dots, x'_{n+1}$, $Y$ is the label vector $(y_1, \dots, y_{n+1})'$, $I$ is the unit $d \times d$ matrix.  
Hence, the vector of nonconformity scores $(R_1, \dots, R_{n+1})'$ can be written as 
$ \lvert Y - H Y \rvert = \lvert (I - H)Y \rvert,$
where $H$ is the hat matrix. \\
Let $y = y_{n+1}$ be a possible label for $x_{n+1}$, and $(z_1, \dots, z_n, (x_{n+1},y))$ the augmented data set. Now, $Y \coloneqq (y_1,\dots, y_n, y)'$. Note that $Y = (y_1,\dots, y_n, 0)' + (0,\dots,0, y)'$ and so the vector of nonconformity scores can be represented as $\lvert A + B y \rvert$, where:
$ A = (I - H) (y_1,\dots, y_n, 0)' $ and
$ B = (I - H) (0,\dots,0, y)'.$
Therefore, each $R_i = R_i(y)$ has a linear dependence on $y$. As a consequence, since the p-value $p_z(y)$ simply counts how many scores $R_i$ are greater than $R_{n+1}$, it can only change at points where $R_i(y) - R_{n+1}(y)$ changes sign for some $i = 1, \dots, n.$ 
This means that we can calculate the set of points $y$ on the real line whose corresponding p-value $p_z(y)$ exceeds $\alpha$ rather than trying all possible $y$, leading to a feasible prediction. 
Precise computations can be found in \cite{book:alrw}, chap 2.

Before going on in the discussion, a clarification is required. 
The reader has surely noticed that the proposed non-conformity measure is computed on the $i$-th example too, contrary to what has been stated before.
In fact, the statement of the conformal algorithm, we define the nonconformity score for the $i$th example by:
$R_i \coloneqq A \, ( \, \Lbag z_1, \dots, z_{i-1}, z_{i+1}, \dots, z_n, z \Rbag \, , z_i)$
(\ref{eq:residualcp}),
apparently specifying that we do not want to include $z_i$ in the bag to which it is compared. But then, in the RRCM, we use the nonconformity scores (\ref{eq:residualrrcm}), as if:
$R_i \coloneqq A \, ( \, \Lbag z_1, \dots, z_n, z \Rbag \, , z_i).$

The point is whether to include the new example in the bag with which we are comparing it or not is a delicate question that is for instance tackled in \cite{paper:tutorial} . 
First of all, it is noteworthy to assert that both of them are valid choices.
They both indeed satisfy the symmetry property required for valid non-conformity measures. In some cases (e.g., when there is a monotonic relation between $A\left(\left\{ z_1,\ldots,z_{i-1},z_i,z_{i+1},z_n,z\right\},z_i\right)$ and $A\left(\left\{ z_1,\ldots,z_{i-1},z_{i+1},z_n,z\right\},z_i\right)$) they can even lead to the same prediction sets. For example, if $R_i$ is the absolute value of the difference between $z_i$ and the mean value of the bag $B$, including or not $z_i$ in the bag is absolute invariant, as in both cases Simple computations show that the two scores are the same, except for a scale factor $\frac{n}{n+1}$.  \newline
From an efficiency viewpoint, one may correctly wonder if $z_i$ should be included in the training set or not in order to maximize the efficiency. To our knowledge there are no works dealing in detail on this issue. Even though the answer could possibly depend on the specific non-conformity measure, underlying model, and data distribution, we believe that this issue could and should be an object of further future investigation.

We have introduced conformal prediction with the formula (\ref{eq:residualcp}), as the reference book of \cite{book:alrw} and the first works did.  
Moreover, in this form conformal prediction generalizes to online compression models (Section 3 of the Supplementary Material \citep{zeni_supp}). 
In general, however, the inclusion of the $i$th example simplifies the implementation or at least the explanation of the conformal algorithm. 
From now on, we rely on this approach when using conformal prediction, and define instead the methods relying on (\ref{eq:residualcp}) as jackknife procedures.

Conformal predictors can be implemented in a feasible and at the same time particularly simple way for nonconformity measures based on the nearest neighbors algorithm, too. 
Recently, an efficient method to compute in an exact way conformal prediction with the Lasso, i.e. considering the quadratic loss function and the $\mathit{l}_1$ norm penalty, has been provided by \cite{lei2019fast}.  A straight extension to the elastic net --- which considers both a $\mathit{l}_1$ and $\mathit{l}_2$ penalty, is also given.

Recently, novel methods to perform regression, that follow a different paradigm to compute NCMs have been proposed: we provide a discussion about them in Section \ref{sec:validity}

\subsection{The Online Framework}\label{sec:online}

The first works on Conformal algorithms tended to focus on the online framework, where examples arrive one by one and so predictions are based on an accumulating data set. 
The predictions these algorithms make 
are \textit{hedged\/}: they incorporate a valid indication of their own accuracy and reliability. 
\cite{book:alrw} claim that most existing algorithms for hedged prediction first learn from a training data set and then predict without ever learning again. The few algorithms that do learn and predict simultaneously, instead, do not provide confidence information.

Moreover, the property of validity of conformal predictors can be stated in an especially strong form in the online framework.
Classically, a method for finding $(1-\alpha)$ prediction regions is considered valid if it has a $(1-\alpha)$ probability of containing the label predicted, because by the law of large numbers it would then be correct $(1-\alpha)$\% of the times when repeatedly applied to independent data sets. 
The online story may seem more complicated, because we repeatedly apply a method not to independent data sets, but to an accumulating data set. 
After using $(x_1,y_1),(x_2,y_2), \dots, (x_n,y_n)$ and $x_{n+1}$ to predict $y_{n+1}$, we use $(x_1,y_1),\dots, (x_{n+1},y_{n+1})$ and $x_{n+2}$ to predict $y_{n+2}$, and so on. 
For a $(1-\alpha)$ online method to be valid, $(1-\alpha)$\% of these predictions must be correct. Under minimal assumptions, conformal prediction is valid in this new and powerful sense.

The intermediate step behind this result is that successive errors are probabilistically independent. 
Let us start with some definitions, following \cite{vovk2002online} and \cite{book:alrw}: being $\omega = (z_1,z_2,\ldots) = (x_1,y_1,x_2,y_2,\ldots$ an infinite data sequence and $\Gamma^\alpha$ a generic randomised conformal predictor, we define the number of errors made by $\Gamma^\alpha$ on the infinite sequence $\omega$ during the first $n$ trials to be
\begin{equation}
    Err_n(\Gamma^\alpha,\omega) := \#\{i=1,\ldots,n: y_i\notin \Gamma^{\alpha}(x_1,y_1,\ldots,x_{i-1},y_{i-1},x_i)\}
\end{equation}
where $\#B$ is the size of set $B$.
The individual prediction result is defined instead as
\begin{equation}
    err_n(\Gamma^\alpha,\omega):=Err_n - Err_{n-1} = \left\{
  \begin{array}{lr}
    1 & y_n\notin \Gamma^{\alpha}(x_1,y_1,\ldots,x_{n-1},y_{n-1},x_n) \\
    0 & otherwise
  \end{array}
\right.
\end{equation}
and the whole infinite sequence of prediction results is then defined as
\begin{equation}
    err(\Gamma^\alpha,\omega) := \left( err_1(\Gamma^\alpha,\omega), err_2(\Gamma^\alpha,\omega),\ldots \right)
\end{equation}

It is shown for the first time in \cite{vovk2002online} that, being and $\Gamma^{\alpha}$ a generic randomised conformal predictor of level $1-\alpha$,  the following holds:
\begin{proposition}\label{prop:onlineindep}
For any randomised conformal predictor $\Gamma$ of any confidence level $1-\alpha$, and any exchangeable probability distribution $P$ in $\boldsymbol{Z}^{\infty}$, the image of $P\times U^{\infty}$ under the mapping
\begin{equation*}
\left(\omega\in \boldsymbol{Z}^{\infty},\tau \in [0,1]^\infty\right) \rightarrow err\left( \Gamma^{\alpha},\omega,\tau\right)   
\end{equation*}
is the probability distribution $B^\infty_\alpha$ of independent Bernoulli random trial of parameter $\alpha.$
\end{proposition}

In a more intuitive sense, Proposition \ref{prop:onlineindep} states, under the exchangeability assumption and in the online mode, that the errors made at different steps by randomised conformal predictors are independent.
The experiment involved in predicting $z_{101}$ from $z_1, \dots, z_{100}$ is not entirely independent of the experiment involved in predicting, say, $z_{105}$ from $z_1, \dots, z_{104}$. 
The $101$ random numbers involved in the first experiment are all also involved in the second one, 
but this overlap does not actually matter. 

Summing up, we already know that the probability of error is below the significance level $\alpha$. 
In addition to that, events for successive $n$ are probabilistically independent notwithstanding the overlap.
Hence, $(1-\alpha)$\% of consecutive predictions must be correct. In other words, the random variables $\mathbbm{1}_{z_{n+1} \notin \gamma^{\alpha} (z_1, \dots, z_{n})}$ are independent Bernoulli variables with parameter $\alpha$.
\cite{vovk2009online} focuses on the prediction of consecutive responses, especially when the number of observations does not exceed the number of parameters.

It should be noted that the assumption of exchangeability rather than randomness makes Proposition~\ref{prop:onlineindep} stronger: it is very easy to give examples of exchangeable distributions on $\mathbf{Z}^N$ that are not of the form $P^N$ --- where it is worth recalling that $P$ is the unknown distribution of examples. 
Nonetheless, in the infinite-horizon case (which is the standard setting for the online mode of prediction) the difference between the exchangeability and randomness assumptions essentially disappears: according to a well-known theorem by de Finetti, each exchangeable probability distribution on $\mathbf{Z}^{\infty}$ is a mixture of power probability distributions $P^{\infty}$, provided $\mathbf{Z}$ is a Borel space \citep{hewitt1955symmetric}. 
In particular, using the assumption of randomness rather than exchangeability in the case of the infinite sequence hardly weakens it: the two forms are equivalent when $\mathbf{Z}$ is a Borel space.



\section{Recent Advances in Conformal Prediction}
\label{part:two}
\subsection{Different Notions of Validity}
\label{sec:validity}

An appealing property of conformal predictors is their automatic validity under the exchangeability assumption: 
\begin{equation}\label{eq:margvalidity}
    \mathbbm{P} (Y_{n+1} \in \Gamma^{\alpha}(Z_1, \dots, Z_n, X_{n+1})) \geq 1 - \alpha \qquad \text{for all } P,
\end{equation}
where $\mathbbm{P} = P^{n+1}$ is the joint measure of $(X_1, Y_1), \dots, (X_{n+1}, Y_{n+1})$. 
A major focus of this section will be on conditional versions of the notion of validity.  

The idea of conditional inference in statistics is about the wish to make conclusions that are as much conditional on the available information as possible. 
Although finite sample coverage defined in (\ref{eq:margvalidity}) is a desirable property, this might not be enough to guarantee good prediction bands, even in very simple cases. 
We refer to (\ref{eq:margvalidity}) as \textit{marginal\/} coverage, which is different from (in fact, weaker than) the \textit{conditional\/} coverage as usually sought in prediction problems. 
As a result, a good estimator must satisfy something more than marginal coverage. A natural criterion would be conditional coverage. 

However, distribution-free \textit{conditional coverage\/}, that is:
\begin{equation}\label{eq:condvalidity}
    \mathbbm{P} (Y_{n+1} \in \Gamma^{\alpha}(x) \mid X_{n+1} = x) \geq 1 - \alpha \qquad \text{for all } P \text{ and a.a }x,
\end{equation}
with $\Gamma^{\alpha}(x) \equiv \Gamma^{\alpha}(Z_1, \dots, Z_n, x)$ 
is impossible to achieve with a finite sample for rich object spaces, such as $\mathbf{X} = \mathbb{R}$ (\cite{paper:lei2014}, Lemma 1). 
Indeed, the requirement of precise object conditional validity cannot be satisfied in a nontrivial way, unless we know the true probability distribution generating the data (or we are willing to use a subjective or postulated probability distribution, as in Bayesian theory), or unless the test object is an atom of the data-generating distribution. 
If we impose that requirement, the prediction interval is expected to have infinite length (\cite{vovk2012conditional} and for general background related to distribution-free inference \cite{bahadur1956nonexistence}, \cite{donoho1988one}).

The negative result --- that conditional coverage cannot be achieved by finite-length prediction intervals without regularity and consistency assumptions on the model and the estimator $f$ --- does not prevent set predictors to be (object) conditionally valid in a partial and asymptotic sense, and simultaneously asymptotically efficient. 

Therefore, as an alternative solution, \cite{paper:lei2014} develop a new notion, called \textit{local validity\/}, that naturally interpolates between marginal and conditional validity, and is achievable in the  finite sample case. 
Formally, given a partition $\mathcal{A} = \{A_j :j \geq 1\}$ of supp$(P_X)$, a prediction band $\Gamma^{\alpha}$ is locally valid with respect to $\mathcal{A}$ if:
\begin{equation}\label{eq:locvalidity}
    \mathbbm{P} (Y_{n+1} \in \Gamma^{\alpha}(X_{n+1}) \mid X_{n+1} \in A_j) \geq 1 - \alpha \qquad \text{for all } j \text{ and all }P.
\end{equation}
Then, their work is focused on defining a method that shows both finite sample (marginal and local) coverage and asymptotic conditional coverage (i.e., when the sample size goes to $\infty$, the prediction band give arbitrarily accurate conditional coverage). At the same time, they prove it to be asymptotic efficient. 
The finite sample marginal and local validity is distribution free: no assumptions on $P$ are required. Then, under mild regularity conditions, local validity implies asymptotically conditionally validity.

The way \cite{paper:lei2014} built the prediction bands to achieve local validity can be seen as a particular case of a bigger class of predictors, which now we introduce and explain, the so called Mondrian conformal predictors.
The discussion about validity is definetely not closed: recently \cite{barber2019limits} reflect again on the idea of a proper intermediate definition, while \cite{romano2019conformalized} proposes the so called Conformalised Quantile Regression (CQR), a novel method that, by fusing a Conformal way of thinking with quantile regression is able to yield valid distribution-free bands, yet remarkably capable to adapt to any heteroschedasticity of the data and with increased efficiency with respect to standard proposals, as thorougly shown by the simulation studies in the article.

As in split conformal prediction, the procedure is based on splitting the data into a proper training set, indexed by $\mathcal{I}_1$ and a calibration set indexed by $\mathcal{I}_2$. Given a quantile regression method $A$, two conditional quantiles $\hat{q}_{\alpha_{hi}}$ and $\hat{q}_{\alpha_{lo}}$ are fitted using the proper training set:
\begin{equation}
    \left\{ \hat{q}_{\alpha_{hi}}, \hat{q}_{\alpha_{lo}}\right\} \longleftarrow A \left(\left\{ (X_i,Y_i: i\in\mathcal{I_1}\right\} \right)
\end{equation}
The non-conformity scores are then calculated as:
\begin{equation}
    E_i := max\left\{\hat{q}_{\alpha_{lo}}(X_i) - Y_i, Y_i - \hat{q}_{\alpha_{hi}}(X_i) - Y_i \right\}
\end{equation}
Finally, given a new input data $X_{n+1}$ the prediction interval for $Y_{n+1}$ is.
\begin{equation}
    C(X_{n+1}) = [\hat{q}_{\alpha_{lo}}(X_{n+1})-Q_{1-\alpha}(E,\mathcal{I}_2),\hat{q}_{\alpha_{hi}}(X_{n+1})+Q_{1-\alpha}(E,\mathcal{I}_2)]
\end{equation}
where $Q_{1-\alpha}(E,\mathcal{I}_2) := (1-\alpha)(1+1/|\mathcal{I}_2|)$-th empirical quantile of ${E_i,i\in \mathcal{I}_2}$
Using the same line of reasoning, an extension of this context to the calculation of conditionally valid confidence (with respect to prediction) intervals for parameters has been very recently proposed in \cite{medarametla2021distributionfree}

\subsubsection{Mondrian Conformal Predictors}\label{sec:mondrian}

We start from an example. 
In handwritten digit recognition problems, some digits (such as ``5'') are more difficult to recognize correctly than other digits (such as ``0''), and it is natural to expect that at the confidence level 95\% the error rate will be significantly greater than 5\% for the difficult digits; our usual, unconditional, notion of validity only ensures that the average error rate over all digits will be close to 5\%. 

We might not be satisfied by the way the conformal predictors work. 
If our set predictor is valid at the significance level 5\% but makes an error with probability 10\% for men and 0\% for women, both men and women can be unhappy with calling 5\% the probability of error.  
It is clear that whenever the size of the training set is sufficient for making conditional claims, we should aim for this. 
The requirement of object conditional validity is a little bit more than what we can ask a predictor to be, but it can be considered as a special case: for somehow important events $E$ we do not want the conditional probability of error given $E$ to be very different from the given significance level $\alpha$.

We are going to deal with a natural division of examples into several \textit{categories\/}: e.g., different categories can correspond to different labels, or kinds of objects, or just be determined by the ordinal number of the example.  
As pointed out in the examples above, conformal predictors --- as we have seen so far --- do not guarantee validity within categories: the fraction of errors can be much larger than the nominal significance level for some categories, if this is compensated by a smaller fraction of errors for other categories. 
A stronger kind of validity, validity within categories, which is especially relevant in the situation of asymmetric classification, is the main property of \textit{Mondrian conformal predictors\/} (MCPs), first introduced in \cite{vovk2003mondrian}.
The exchangeable framework is the assumption under which MCPs are proved to be valid; in Section 3 of the Supplementary Material \citep{zeni_supp}, again, we will have a more general setting, relaxing the hypothesis. 

When the term categories comes into play, we are referring to a given division of the example space $\mathbf{Z}$: a measurable function $\kappa$ maps each $z$ to its category $k$, belonging to the (usually finite) measurable space $\mathbf{K}$ of all categories. In many instances, it is a kind of classification of $z_i$. 
The category $\kappa_i = \kappa(z_i)$ might depend on the other examples in the data sequence $(z_1, \dots, z_n),$ but disregarding their order. 
Such a function $\kappa$ is called a Mondrian taxonomy, as a tribute to the Dutch painter Piet Mondrian. Indeed, the taxonomy that $\kappa$ defines in the space $\mathbf{Z}$ recalls the grid-based paintings and the style for which the artist is renowned. 

To underline the dependence of $\kappa(z_i)$ on the bag of the entire dataset, \cite{book:2014} introduce the $n$-taxonomy $K : \mathbf{Z}^n \Rightarrow \mathbf{K}^n$, which maps a vector of examples to the vector of corresponding categories. 
Using this notation, it is required that the $n$-taxonomy $K$ is equivariant with respect to permutations, that is:  
\[ (\kappa_1, \dots, \kappa_n) = K(z_1, \dots, z_n) \Rightarrow (\kappa_{\pi(1)}, \dots, \kappa_{\pi(n)}) =K(z_{\pi(1)}, \dots, z_{\pi(n)}) .\] 
We prefer however to let the dependence implicit and remain stuck to the simpler notation of \cite{book:alrw}.

Given a Mondrian taxonomy $\kappa$, to use conformal prediction we have to modify slightly some of the definitions seen in the previous chapter. 
To be precise, a Mondrian nonconformity measure might take into account also the categories $\kappa_, \dots, \kappa_n$, while the p-values (\ref{eq:pvalue}) should be computed as:
\begin{equation}
 p_z \coloneqq \dfrac{ \lvert \{ \, i = 1, \dots, n+1 : \kappa_i = \kappa_{n+1} \,\&\, R_i \geq R_{n+1}\} \rvert }{\lvert \{ \, i = 1,\ldots,n + 1 : \kappa_i = \kappa_{n+1}  \} \rvert },
\end{equation}
where $\kappa_{n+1} = \kappa(z)$. 
As a remark, we would like to point out and stress what we are exactly doing in the formula just defined. Although one can choose any conformity measure, in order to have local validity the ranking must be based on a local subset of the sample.
Hence, the algorithm selects only the examples among the past experience that have the same category of the new one, and makes its decision based on them.

At this point, the reader is able to write by himself the smoothed version of the MCP, which satisfies the required level of reliability in an exact way. 
\begin{proposition}
If examples $z_1, \dots, z_{n+1}$ are generated from an exchangeable probability distribution on $\mathbf{Z}^{n+1}$, any smoothed MCP based on a Mondrian taxonomy $\kappa$ is category-wise exact with respect to $\kappa$.
\end{proposition} 
In the previous proposition,category-wise exact means that  $\forall n$, the probability of $p_{n+1}$ conditional to $k(1,z_1),p_1,\ldots,k(n,z_n),p_n$ is a uniform on $[0,1]$ where $z_1,\ldots,z_n,z_{n+1}$ are generated from an exchangeable distribution $Z^{\infty}$ and $p_1,\ldots,p_n$ are the $p-$values generated by $f$ \citep[][Section 4.5]{book:alrw}.

Moreover, we might want to have different significance levels $\alpha_k$ for different categories $k$. 
In some contexts, certain kinds of errors are more costly than others. For example, it may be more costly to classify a high-risk credit applicant as low risk (one kind of error) than it is to classify a low-risk applicant as high risk (a different kind of error). 
In an analogous way, we could be required to distinguish between useful messages and spam in the problem of mail filtering: classifying a useful message as spam is a more serious error than vice versa. 
We do not have misclassification costs to take into account, but setting in a proper way the significance levels allow us to specify the relative importance of different kinds of prediction errors. And MCPs still do the job \citep{book:alrw}.

Last, a brief discussion of an important question: how to select a good taxonomy? While choosing the partitions that determine a Mondrian taxonomy $\kappa$, it comes out indeed a dilemma that is often called the ``problem of the reference class''. 
We want the categories into which we divide the examples to be large, in order to have a reasonable sample size for estimating the probabilities. But we also want
them to be small and homogeneous, to make the inferences as specific as possible. 
\cite{book:2014} points out a possible strategy for conditional conformal predictors in the problem of classification in the online setting.  
The idea is to adapt the method as the process goes on. At first, the conformal predictor should not be conditional at all. Then, as the number of examples grows, it should be label conditional. 
As the number of examples grows further, we could split the objects into clusters (using a label independent taxonomy) and make the prediction sets conditional on them as well.

\subsection{Inductive Prediction}
\label{sec:inductive}

While being very straightforward from a mathematical point of view, Transductive (or "Full") Conformal predictors are very computationally intensive and, over time, an extensive literature has developed to address this issue.
In particular, \textit{inductive conformal predictors\/} (ICPs), also referred as \textit{split conformal predictors\/}  have been proposed 

ICPs were first proposed by \cite{papadopoulos2002inductive} for regression and by \cite{papadopoulos2002qualified} for classification, and in the online setting by \cite{vovk2002online}. 
Before the appearance of inductive conformal predictors, several other possibilities had been studied, but not with great success. 
To speed computations up in a multi-class pattern recognition problem which uses support vector machines in its implementation, \cite{saunders2000computationally} used a hashing function to split the training set into smaller subsets, of roughly equal size, which are then used to construct a number of support vector machines. 
In a different way, just to mention but a few, \cite{ho2004learning} exploit the adiabatic version of incremental support vector machine, and lately \cite{vovk2013transductive} introduces Bonferroni predictors, a simple modification based on the idea of the Bonferroni adjustment of p-values.

\begin{figure}
    \centering
    \includegraphics[width=9cm]{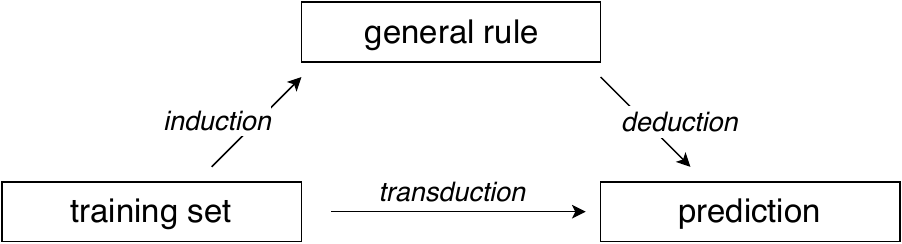}
    \caption[Inductive and transductive approach to prediction.] {Inductive and transductive approach to prediction.}
    \label{fig:induction}
\end{figure}
We now spend some words to recall the concepts of transduction and induction (figure~\ref{fig:induction}), as introduced in \cite{vapnik1998statistical}. 
In inductive prediction we first move from the training data to some general rule: a prediction or decision rule, a model, or a theory (inductive step). When a new object comes out, we derive a prediction based on the general rule (deductive step). 
On the contrary, in transductive prediction, we take a shortcut, going directly from the old examples to the prediction for the new object.  
The practical distinction between them is whether we extract the general rule or not. A side-effect of using a transductive method is computational inefficiency; computations need to be started from scratch every time.

Combining the inductive approach with conformal prediction, the data sequence \newline $(z_1, \dots, z_n)$ is \textit{split\/} into two parts, the proper training set $(z_1, \dots, z_m)$ of size $m < n$ and the \textit{calibration set\/} $(z_{m+1}, \dots, z_n).$ 
We use the proper training set to feed the underlying algorithm, and, using the derived rule, we compute the non-conformity scores for each example in the calibration set. 
For every potential label $y$ of the new unlabelled object $x_{n+1}$, its score $R_{n+1}$ is calculated and is compared to the ones of the calibration set.
In more mathematical terms, we define the nonconformity scores as:
\begin{equation}
    \begin{split}
        R_j & \coloneqq A_{m_k + 1}\left(\Lbag (x_1,y_1),\ldots,(x_{m_k},y_{m_k}) \Rbag, (x_j,y_j)\right),\, for\, j=m_k + 1,\ldots,n-1 \\
        R_{n+1} & \coloneqq A_{m_k + 1}\left(\Lbag (x_1,y_1),\ldots,(x_{m_k},y_{m_k}) \Rbag, (x_{n+1},y) \right), \\
    \end{split}
\end{equation}
Therefore the p-value is:
\begin{equation}
    \label{eq:pvalueinduct}
    p_z \coloneqq \dfrac{ \lvert \{ \, i = m+1, \dots, n+1 : R_i \geq R_{n+1}\} \rvert }{n-m+1}.
\end{equation}
Inductive conformal predictors can be smoothed in exactly the same way as conformal predictors.
As in the transductive approach, under the exchangeability assumption, $p_z$ is a valid p-value. All is working as before.
For a discussion of conditional validity and various ways to achieve it using inductive conformal predictors, see \cite{vovk2012conditional}. 

A greater computational efficiency of inductive conformal predictors is now evident. 
The computational overhead of ICPs is light: they are almost as efficient as the underlying algorithm. The decision rule is computed from the proper training set only once, and it is applied to the calibration set also only once. 
Several studies related to this fact are reported in the literature. For instance, a computational complexity analysis can be found in the work of \cite{papadopoulos2008inductive}, where conformal prediction on top of neural networks for classification has been closely examined.

With such a dramatically reduced computation cost, it is possible to combine easily conformal algorithms with computationally heavy estimators. 
While validity is taken for granted in conformal framework, efficiency is related to the underlying algorithm.
Taking advantage of the bargain ICPs represent, we can compensate the savings in computational terms and, in metaphor, invest a lot of resources in the choice of $f$.

Moreover, this computational effectiveness can be exploited further and fix conformal prediction as a tool in Big Data frameworks, where the increasing size of datasets represents a challenge for machine learning and statistics. 
The inductive approach makes the task feasible, but can we ask for anything more? Actually, the (trivially parallelizable) serial code might be run on multiple CPUs.  
\cite{capuccini2015conformal} propose and analyze a parallel implementation of the conformal algorithm, where multiple processors are employed simultaneously in the Apache Spark framework. 

Achieving computational efficiency does not come for free. A drawback of inductive conformal predictors is their potential prediction inefficiency. In actual fact, we waste the calibration set when developing the prediction rule $f$, and we do not use the proper training set when computing the p-values. 
An interesting attempt to cure this disadvantage is made in \cite{vovk2015cross}. 
\textit{Cross-conformal prediction\/}, a hybrid of the methods of inductive conformal prediction and cross-validation, consists, in a nutshell, in dividing the data sequence into $K$ folds, constructing a separate ICP using the $k$th fold as the calibration set and the rest of the training set as the proper training set. 
Then the different p-values, which are the outcome of the procedure, are merged in a proper way.

Of course, it is also possible to use a uneven split, using a larger portion of data for model fitting and a smaller set for the inference step. 
This will produce sharper prediction intervals, but the method will have higher variance; this trade-off is unavoidable for data splitting methods. 
Common choices found in the applied literature for the dimension of the calibration set, providing a good balance between underlying model performance and calibration accuracy, lie between $25\%$ and $33\%$ of the dataset. 
The problem related to how many examples the calibration set should contain is faced meticulously in \cite{linusson2014efficiency}. 
To maximize the efficiency of inductive conformal classifiers, they suggest to keep it small relative to the amount of available data (approximately $15\% - 30\%$ of the total). At the same time, at least a few hundred examples should be used for calibration (to make it granular enough), unless this leaves too few examples in the proper training set.
Techniques that try to handle the problems associated with small calibration sets are suggested and evaluated in both \cite{johansson2015handling} and \cite{carlsson2015modifications}, using interpolation of calibration instances and a different notion of (approximate) p-value, respectively.

Splitting improves dramatically on the speed of conformal inference, but it introduces additional noise into the procedure. 
One way to reduce this extra randomness is to combine inferences from $N$ several splits, each of them --- using a Bonferroni-type argument --- built at level $1 - \alpha / N$ . 
Multiple splitting on one hand decreases the variability as expected, but on the other hand this may produce, as a side effect, the width of $\Gamma_N^{\alpha}$ to grow with $N$.
As described in \cite{paper:tutorial}, under rather general conditions, the Bonferroni effect is dominant and hence intervals get larger and larger with $N$. For this reason, they suggest using a single split. 

\cite{linusson2014efficiency} even raise doubts about the commonly accepted claim that transductive conformal predictors are by default more efficient than inductive ones.
It is known indeed that an unstable nonconformity function --- one that is heavily influenced by an outlier example, e.g., an erroneously labeled new example $(x_{n+1},y)$ --- can cause (transductive) conformal confidence predictors to become inefficient. 
They compare the efficiency of transductive and inductive conformal classifiers using decision tree, random forest and support vector machine models as the underlying algorithm, to find out that the full approach is not always the most efficient. 
Their position is actually the same of \cite{papadopoulos2008inductive}, where the loss of accuracy introduced by induction is claimed to be small, and usually negligible. And not only for large data sets, which clearly contain enough training examples so that the removal of the calibration examples does not make any difference to the training of the algorithm.
Some further explorations on the topic can be found in \cite{linusson2017calibration} but, to our knowledge, the question remains still open.

From another perspective, lying between the computational complexities of the full and split conformal methods is \textit{jackknife prediction\/}. 
This method wish to make a better use of the training data than the split approach does and to cure as much as possible the connected loss of informational efficiency, when constructing the absolute residuals, due to the partition of old examples into two parts, without resorting at the same time to the extensive computations of the full conformal prediction. 
With this intention, it uses leave-one-out residuals to define prediction intervals. 
That is to say, for each example $z_i$ it trains a model $f_{-i}$ on the rest of the data sequence $(z_1, \dots, z_{i-1}, z_{i+1}, \dots, z_n)$ and computes the nonconformity score $R_i$ with respect to $f_{-i}.$

The advantage of the jackknife method over the split conformal method is that it can often produce regions of smaller size. 
However, in regression problems it is not guaranteed to have valid coverage in finite samples. As \cite{paper:lei2017} observe, the jackknife method has the finite sample in-sample coverage property:
\begin{equation}
    \mathbb{P}\, (Y_i \in \Gamma_{jack}^{\alpha}(X_i)) \geq 1 - \alpha,
\end{equation}
where $\Gamma_{jack}^{\alpha}(X_i)$ is the Jackknife prediction set. 
When dealing with out-of-sample coverage (actually, true predictive inference), the properties of the Jackknife are much more fragile.
In fact, even asymptotically, its coverage properties do not hold without requiring nontrivial conditions on the base estimator $f$. 
It is actually due to the approximation required to avoid the unfeasible enumeration approach, that we are going to tackle in a while, precisely in the next section. 
The predictive accuracy of the jackknife under assumptions of algorithm stability is explored by \cite{steinberger2016leave} for the linear regression
setting, and in a more general setting by \cite{steinberger2018conditional}.
Hence, while the full and split conformal intervals are valid under essentially no assumptions, the same is not true for the jackknife ones.

The key to speed up the learning process is to employ a fast and accurate learning method as the underlying algorithm. 
This is exactly what \cite{wang2018fast} do, proposing a novel, fast and efficient conformal regressor, with combines the local-weighted (see Section~\ref{sec:other}) jackknife prediction, and the regularized extreme learning machine.
Extreme learning machine (ELM) addresses the task of training feed-forward neural networks fast without losing learning ability and predicting performance. 
The underlying learning process and the outstanding learning ability of ELM make the conformal regressor very fast and informationally efficient.

Recently, a slight but crucial and remarkable modification to the algorithm gives life to the jackknife+ methods, able to restore rigorous coverage guarantees \citep{barber2020predictive}. We expect many possible extension of these methods to stem from this seminal piece of research.


\subsection{Other Interesting Developments}\label{sec:other}

Full conformal and split conformal methods, combined with basically any fitting procedure in regression, provide finite sample distribution-free predictive inference.
We are now going to introduce generalizations and further explorations of the possibilities of CP along different directions.

In the pure online setting, we get an immediate feedback (the true label) for every example that we predict. 
While this scenario is convenient for theoretical studies, in practice, however, rarely one immediately gets the true label for every object. 
On the contrary weak teachers are allowed to provide the true label with a delay or sometimes not to provide it at all. 
In this case, we have to accept a weaker (actually, an asymptotic) notion of validity, but conformal confidence predictors adapt and keep at it \citep{ryabko2003online,nouretdinov2006criterion}.

Moreover, we may want something more than just providing p-values associated with the various labels to which a new observation could belong. 
We might be interested in the problem of probability forecasting: we observe $n$ pairs of objects and labels, and after observing the $(n+1)$th object $x_{n+1}$, the goal is to give a probability distribution $p_{n+1}$ for its label. 
It represents clearly a more challenging task (\cite{book:alrw}, chap 5), therefore a suitable method is necessary to handle carefully the reliability-resolution trade-off. 
A class of algorithms called \textit{Venn predictors\/} \citep{vovk2004self} satisfies the criterion for validity when the label space is finite, while only among recent developments there are adaptations in the context of regression, i.e. with continuous labels --- namely \cite{nouretdinov2018inductive} and in a different way, following the work of \cite{shen2018prediction}, \cite{vovk2017nonparametric}.
For many underlying algorithms, Venn predictors (like conformal methods in general) are computationally inefficient. Therefore \cite{lambrou2012reliable}, and as an extension \cite{lambrou2015inductive}, combine Venn predictors and the inductive approach, while \cite{vovk2018cross} introduce cross-conformal predictive systems. 

Online compression models is not the only framework where CP does not require examples to be exchangeable.
\cite{dunn2018distribution} extend the conformal method to construct valid distribution-free prediction sets when there are random effects, and \cite{barber2019conformal} to handle weighted exchangeable data, as in the setting of covariate shift \citep{shimodaira2000improving, chen2016robust}. 
\cite{dashevskiy2011time} robustify the conformal inference method by extending its validity to settings with dependent data. They indeed propose an interesting blocking procedure for times series data, whose theoretical performance guarantees are provided in \cite{chernozhukov2018exact}.

Now, we describe more in detail a couple of other recent advances, while advances on High-Dimensional Regression in a conformal setting are described in Section 4 of the Supplementary Material \citep{zeni_supp}.

\subsubsection{Normalized Nonconformity Scores}\label{subsecOD:normaliz}

\begin{figure}
\centering
\includegraphics[width=12cm]{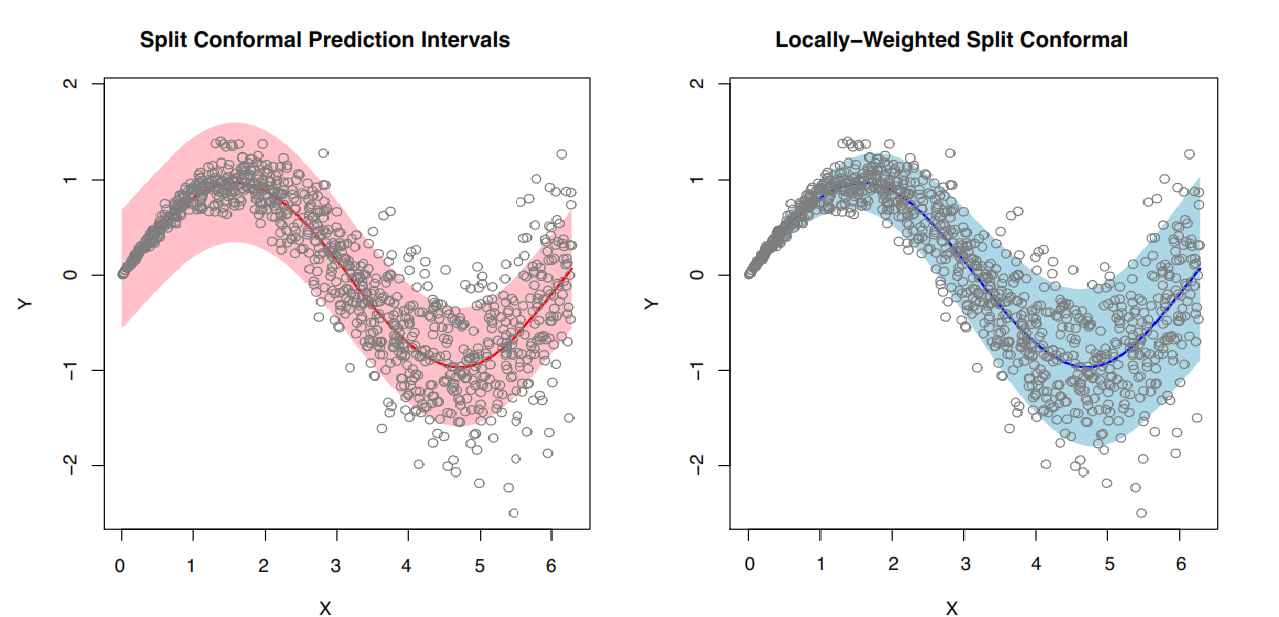}
\caption[Local-weighted conformal inference. \textit{Source: \cite{paper:lei2017}}.]
{Conformal predictors do not contemplate heteroskedasticity in the data distribution. In such a case, one would expect the length of the output interval to be an increasing function of the corresponding variance of the output value, which can give more information of the target label. 
To tackle this problem, local-weighted conformal inference has been introduced. 
\textit{Source: \cite{paper:lei2017}}.}
\label{fig:locally}
\end{figure}

In conformal algorithms seen so far, the width of $\Gamma^{\alpha}(x)$ is roughly immune to $x$ (figure~\ref{fig:locally}, left). 
This property is desirable if the spread of the residual $Y - \hat{Y}$, where $\hat{Y} = f(X)$, does not vary substantially as $X$ varies. 
However, in some scenarios this will not be true, and we wish conformal bands to adapt correspondingly. 
Actually, it is possible to have individual bounds for the new example which take into account the difficulty of predicting a certain $y_{n+1}$. 
The rationale for this, from a conformal prediction standpoint, is that if two examples have the same nonconformity scores using (\ref{eq:residgeneral}), but one is expected to be more accurate than the other, then the former is actually stranger (more nonconforming) than the latter. 
We are interested in resulting prediction intervals that are smaller for objects that are deemed easy to predict and larger for harder objects. 

To reach the goal, normalized nonconformity functions come into play (figure~\ref{fig:locally}, right), that is:
\begin{equation}\label{eq:residualnormal}
    R_i = \frac{\lvert \hat{y}_i - y_i \rvert}{\sigma_i},
\end{equation}
where the absolute error concerning the $i$th example is scaled using the expected accuracy $\sigma_i$ of the underlying model; see, e.g., \cite{papadopoulos2011reliable}, and \cite{papadopoulos2011regression}.
Choosing Equation (\ref{eq:residualnormal}), the confidence predictor (Equation 2.1 in the Supplementary Material \citep{zeni_supp}) becomes: 
\begin{equation}
\Gamma^{\alpha}(x_{n+1}) = [f(x_{n+1}) - R_s \, \sigma_{n+1}, f(x_{n+1}) + R_s \, \sigma_{n+1}]. 
\end{equation}
As a consequence, the resulting predictive regions tend to be tighter than those produced by the simple conformal methods. 
Anyway, using locally-weighted residuals, as in (\ref{eq:residualnormal}), the validity and accuracy properties of the conformal methods, both finite sample and asymptotic, again carry over. 

As said, $\sigma_i$ is an estimate of the difficulty of predicting the label $y_i$. There is a wide choice of estimates of the accuracy available in the literature. A common practice is to train another model to predict errors, as in \cite{papadopoulos2010neural}.
More in details, once $f$ has been trained and the residual errors computed, a different model $g$ is fit using the object $x_1, \dots, x_n$ and the residuals. Then, $\sigma_i$ could be set equal to $g(x_i) + \beta$, where $\beta$ is a sensitivity parameter that regulates the impact of normalization. 

Other approaches use, in a more direct way, properties of the underlying model $f$; for instance, it is the case of \cite{papadopoulos2008normalized}. 
In the paper, they consider conformal prediction with $k$-NNR method, which computes the weighted average of the $k$ nearest examples, and as a measure of expected accuracy they simply use the distance of the examined example from its nearest neighbours. Namely, 
\begin{equation}\label{eq:normalaccuracy}
   d_i^k = \sum_{j=1}^k \text{distance}\,(x_i, x_j).
\end{equation}
The nearer an example is to its neighbours, the more accurate this prediction is indeed expected to be. 


\subsubsection{Functional Prediction Bands}\label{subsecOD:func}

Functional Data Analysis (FDA) is a branch of statistics that analyses data that exist over a continuous domain, broadly speaking functions. 
Functional data are intrinsically infinite dimensional. This is a rich source of information, which brings many opportunities for research and data analysis --- a powerful modeling tool. 
Meanwhile the high or infinite dimensional structure of the data, however, poses challenges both for theory and computations. 
Therefore, FDA has been the focus of much research efforts in the statistics and machine learning community in the last decade.

There are few publications in the conformal prediction literature that deal with functional data. We are going to give just some details about a simple scenario that could be reasonably typical. In the following, the work of \cite{lei2015conformal} guide us. 
The sequence $z_1(\cdot), \dots, z_n(\cdot)$ consists now of $L^2[0,1]$ functions.  
The definition of validity for a confidence predictor $\gamma^{\alpha}$ is: 
\begin{equation}\label{eq:functvalidity}
    \mathbbm{P} \left(z_{n+1}(t) \in \gamma^{\alpha}(z_1, \dots, z_n)(t) \, \, \forall t \, \right) \geq 1 - \alpha \qquad \text{for all } P.
\end{equation}
Then, as always, to apply conformal prediction, a nonconformity measure is needed. A fair choice might be: 
\begin{equation}
R_i = \int (z_i (t) - \bar{z}(t))^2 dt,
\end{equation}
where $\bar{z}(t)$ is the average of the augmented data set. 
Due to the dimension of the problem, an inductive approach is more desirable. Therefore, once the nonconformity scores $R_i$ are computed for the example functions of the calibration set, the conformal prediction set is given by all the functions $z$ whose score is smaller than the suitable quantile $R_s$. \\
Then, one more step is mandatory. Given a conformal prediction set $\gamma^{\alpha}$, the inherent prediction bands are defined in terms of lower and upper bounds: 
\begin{equation}
l(t) = \inf_{z \in \gamma^{\alpha}} z(t) \qquad \text{and} \qquad u(t) = \sup_{z \in \gamma^{\alpha}} z(t).
\end{equation}
Consequently, thanks to provable conformal properties, 
\begin{equation}\label{eq:funcbands}
    \mathbbm{P} \left(l(t) \leq z_{n+1}(t) \leq u(t), \, \, \forall t\, \right) \geq 1 - \alpha.
\end{equation}
However, $\gamma^{\alpha}$ could contain very disparate elements, hence no close form for $l(t)$ and $u(t)$ is available in general and these bounds may be hard to compute.

To sum up, the key features to be able to handle functional data efficiently are the nonconformity measure and a proper way to make use of the prediction set in order to extract useful information. 
The question is still an open challenge, but the topic stands out as a natural way for conformal prediction to grow up and face bigger problems.

An intermediate work in this sense is \cite{lei2015conformal}, which studies prediction and visualization of functional data paying specific attention to finite sample guarantees.
As far as we know, it is the only analysis up to now that applies conformal prediction to the functional setting. 
In particular, their focal point is exploratory analysis, exploiting conformal techniques to compute clustering trees and simultaneous prediction bands --- that is, for a given level of confidence $1-\alpha$, the bands that covers a random curve drawn from the underlying process (as in \ref{eq:functvalidity}).

However, satisfying (this formulation of) validity could be really a tough task in the functional setting. 
Since their focus is on the main structural features of the curve, they lower the bar and set the concept in a revised form, that is:
\begin{equation}\label{eq:functnewval}
    \mathbbm{P} (\,\Pi(z_{n+1}) (t) \in \gamma^{\alpha}(z_1, \dots, z_n)(t)\, \, \forall t\,) \geq 1 - \alpha \qquad \text{for all } P,
\end{equation}
where $\Pi$ is a mapping into a finite dimensional function space $\Omega_p \subseteq L^2[0,1]$. 

The prediction bands they propose are constructed, as shown previously in (\ref{eq:functnewval}), adopting a finite dimensional projection approach. 
Once a basis of functions $\{\phi_1, \dots, \phi_p \}$ is chosen --- let it be a fixed one, like the Fourier basis, or a data-driven basis, such as functional principal components --- the vector of projection coefficients $\xi_i$ is computed for each of the $m$ examples in the proper training set. 
Then, the scores $R_i$ measure how different the projection coefficients are with respect to the ones of the training set, that is, for the $i$th calibration example, $R_i = A(\xi_1, \dots, \xi_{m}; \xi_i)$. 
Let:
\begin{equation}
\gamma_{\xi} = \{ \xi \in \mathbb{R}^p : R_{\xi} \leq R_s \}
\end{equation} and 
\begin{equation}
    \gamma^{\alpha}(t) = \left\{ \sum_{i=1}^p \varsigma_i \phi_i(t) : (\varsigma_1, \dots, \varsigma_p) \in \gamma_{\xi} \right\}.
\end{equation}
As a consequence, $\gamma^{\alpha}$ is valid, i.e. (\ref{eq:functnewval}) holds.

Exploiting the finite dimensional projection, the conformity measure handles vectors, so all the experience seen in these two chapters gives a hand. A density estimator indeed is usually selected to assess conformity.
Nevertheless, picking $A$ out is critical in the sense that a not suitable one may give a lot of trouble in computing $\gamma^{\alpha}(t)$. It is the case, for instance, of kernel density estimators.
In their work, the first $p$ elements of the eigenbasis --- i.e. the eigenfunctions of the autocovariance operator --- constitute the basis, while $A$ is a Gaussian mixture density estimator. In this set up, approximations are available, and lead to the results they obtain.

Though their work can be deemed as remarkable, the way used to proceed simplifies a lot the scenario. It is a step forward in order to extend conformal prediction to functional data, but not a complete solution.  
Very promising results in extending and overcoming the approach in \cite{lei2015conformal}, can be found in \cite{diquigiovanni2021importance}, where the authors propose a family of novel conformal predictors for functional data that yield closed-form simultaneous and finite-sample valid prediction bands, as well as proposing an approach to modulate the amplitude of the bands in relationship to the problem at hand.
The same authors moreover propose an extension of such approach to a multivariate functional data setting and in a regression framework in \cite{diquigiovanni2021conformal}. Moving from \cite{chernozhukov2018exact} it is also proposed, in \cite{diquigiovanni2021distribution}, an extension of the proposed method to a functional time-series setting, with an application to the modelling of the gas market.

\subsection*{Acknowledgments}
The authors acknowledge financial support from: Accordo Quadro ASI-POLIMI ``Attivit\`a di Ricerca e Innovazione'' n. 2018-5-HH.0, collaboration agreement between the Italian Space Agency and Politecnico di Milano; the European Research Council, ERC grant agreement no 336155-project COBHAM ``The role of consumer behaviour and heterogeneity in the integrated assessment of energy and climate policies''; the ``Safari Njema Project - From informal mobility to mobility policies through big data analysis'', funded by Polisocial Award 2018 - Politecnico di Milano.
The authors would also like to thank two anonymous reviewers for the invaluable suggestions provided.

\begin{supplement}

\stitle{Supplementary Material to "Conformal Prediction: a Unified Review of Theory and New Challenges"}

\sdescription{Additional Sections and Miscellaneous Topics}

\end{supplement}


\end{document}